\documentclass[runningheads]{llncs}
\pdfoutput=1
\usepackage[utf8]{inputenc}
\usepackage[english]{babel}

\usepackage{float}
\usepackage{amsmath}
\usepackage{amssymb}
\usepackage{algpseudocode}
\usepackage{algorithm, algorithmicx}
\usepackage{subcaption}
\usepackage{adjustbox}
\usepackage{hhline}
\usepackage{graphicx}
\graphicspath{{./images/}}

\usepackage{xcolor}

\newcommand{\entitySet}{\mathcal{E}}
\newcommand{\predSet}{\mathcal{P}}
\newcommand{\timeSet}{\mathcal{T}}

\newcommand{\real}{\mathbb{R}}

\newcommand{\TKG}{\mathit{T}}

\newcommand{\splime}{\textsc{SpliMe}}

\newcommand*\Let[2]{\State #1 $\gets$ #2}

\algtext*{EndFunction}

\AtBeginDocument{}

\title{Leveraging Static Models for Link Prediction in Temporal Knowledge Graphs}

\author{
    Radstok, Wessel \\
	\texttt{w.radstok@gmail.com} \\
    \and
    Chekol, Mel \\ 
	\texttt{m.w.chekol@uu.nl}
}
\institute{Utrecht University}

\date{April 2021}

\begin{document}

\maketitle

\begin{abstract}    
    The inclusion of temporal scopes of facts in knowledge graph embedding (KGE) presents significant opportunities for improving the resulting embeddings, and consequently for increased performance in downstream applications.
    Yet, little research effort has focussed on this area and much of the carried out research reports only marginally improved results compared to models trained without temporal scopes (static models).
    Furthermore, rather than leveraging existing work on static models, they introduce new models specific to temporal knowledge graphs.
    We propose a novel perspective that takes advantage of the power of existing static embedding models by focussing effort on manipulating the data instead.
    Our method, \splime , draws inspiration from the field of signal processing and early work in graph embedding. 
    We show that \splime\ competes with or outperforms the current state of the art in temporal KGE. 
    Additionally, we uncover issues with the procedure currently used to assess the performance of static models on temporal graphs and introduce two ways to counteract them.
%
\end{abstract}

\section{Introduction}
A knowledge graph (KG) is a graph used for representing structured information about the world, which can then be utilized in tasks such as question answering \cite{10.1145/3289600.3290956}, relation extraction \cite{bastos2021recon} and recommender systems \cite{guo2020survey}. Information is stored in the form of (subject, predicate, object) triples, e.g., (\textit{Obama, presidentOf, USA}), also called facts. The subject and object are also referred to as entities. In the graph representation, the entities are nodes and predicates are the labelled directed edges between them.

Knowledge graph completion is the problem of inferring missing facts in a KG.
This is often achieved through KG \emph{embedding} where the entities and predicates making up the KG are embedded into a low-dimensional vector space. 
The embedding models implement scoring functions  to calibrate the positions of entities and predicates in the vector space.
A number of different scoring functions have been proposed in order to successfully capture the underlying structure of a \textit{static} KG.
Surveys of these models can be found in \cite{ji2020survey,nickel2015review}.

A largely unexplored component of KG embedding is the inclusion of the temporal scopes of facts--the time at which a fact occurred or held true. Referring back to our earlier example: Obama was only president of the USA from 2009 to 2017. Alternatively, there exist event KGs where each event is associated with a timestamp (e.g.\textit{ (Obama, awarded, noble-peace-prize, 2015)}). Both are collectively referred to as temporal knowledge graphs (TKG).

It is known that including temporal scopes of facts in a KG embedding would result in improved performance \cite{dasgupta2018hyte,goel2020diachronic,xu2019temporal}. However, comparatively little research has been performed in this area. The performance results of which are only marginally better than that reported by models without temporal scopes. Furthermore, these approaches focus on introducing new models specific to temporal graphs rather than finding ways to leverage the impressive performance of existing static models. We believe that the option of transforming the data (e.g., through splitting) to achieve this has been overlooked. 

We propose a new method called \splime\  which operates on the graph in order to leverage the power of static embedding models. Specifically, \splime\ uses the valid times (temporal scopes) of facts to embed predicates in three different ways: (i) timestamping,  (ii) splitting (leverages change point detection~\cite{aminikhanghahi2017survey}), and (iii) merging. Similarly to~\cite{goel2020diachronic}, \splime\ is model agnostic, meaning it can be used with any existing static embedding model and therefore take advantage of further development in this area.
The core of \splime\ is a transformation function which maps a TKG into a new representation where the temporal scope is included at the level of entities and/or predicates. The resulting KGs can subsequently be embedded using  models for static KGs. We carried out several experiments and compared \splime\ with state-of-the-art TKG embedding models. Despite only performing experiments with TransE, a conceptually simple model, results show that \splime\ outperforms most of these models. We believe that the results can be increased by using more expressive static embedding models such as ComplEx~\cite{pmlr-v48-trouillon16} or SimplE~\cite{NIPS2018_7682}.

Additionally, we uncover issues with the procedure used to evaluate static models on temporal knowledge graphs. Specifically, the current procedure causes test-leakage, i.e., elements from the test set appear in the training set. Due to this, the performance of static models in this scenario is sometimes overestimated. Consequently, the relative increase in performance that temporal models provide is underestimated. We provide two procedures that fix this error and show even with this fix, \splime\ still matches or outperforms its competitors.


\section{Preliminaries}\label{sec:background}
Given an entity set $\entitySet = \{e_1,e_2,\dots,e_n \}$ and a predicate set $\predSet = \{r_1,r_2,\dots,r_m\}$, a knowledge graph $G$  is a set of observed triples, i.e., $G = \{(s,p,o) \ | \ s \in \entitySet,\ p \in \predSet,\ o \in \entitySet \}$. In the graph representation, entities are nodes and predicates are labelled directed edges between them.
Temporal knowledge graphs (TKGs) also include a temporal scope for (a subset of)) facts. Time in this case is modeled as a set of discrete, linearly ordered timestamps: $\timeSet = \{t_1, t_2, \dots, t_l\}$. TKGs come in two varieties: the first is as a set of quadruples, i.e. $\mathit{T} = \{(s,p,o,h) | h \in \timeSet\}$ where $h$ is a timestamp to denote when the fact happened. This method is useful for modelling the occurrences of events. A TKG defined in this manner is also called an \textit{event} TKG. The second variety is as a set of quintuples, i.e., $T = \{(s,p,o,b,e) \ | \ b, e \in \timeSet, b \leq e\}$. This formulation models the time period temporal scope during which facts held true, also known as \emph{valid time} TKG. 

We introduce some notations that will be used throughout the paper. We consider the subgraph $T^{p = r}$ to denote all facts in a TKG containing the predicate $r \in \predSet$, i.e., $T^{p = r}=\{(s,p,o,b,e) \in T \ | \ p = r \}$. Analogously, we define $T^{s = z}$ as all facts containing entity $z \in \entitySet$, $T^{b = t}$ as all facts with start time $t \in \timeSet$ and so on. This indexation can be chained: $T^{p = r, e=t}$ is the TKG containing all facts which contain predicate $r$ and end at timestamp $t$.

\paragraph{Knowledge graph embedding} is the representation of a KG in a continuous, low-dimensional vector space by learning a vector representation for each entity and predicate. 
The vectors represent the \emph{latent features} of entities and relationships: the underlying parameters that determine their interactions. 
For a triple ($s,p,o$), let ($e_s, e_p, e_o$) denote its embedding vectors. Taking a KG and a random initialization of the vectors as an input, a vector representation of the KG is gradually learned using a scoring function $\phi(s,p,o)$. The scoring function should reflect how well the embedding captures the semantics of the KG. For instance, the scoring function of TransE, as explained in further detail in Section~\ref{representationlearning}, is written as $\phi(s,p,o)= ||e_s + e_p - e_o||_{1,2}$. The learned embeddings can be used in tasks such as classification, clustering, and link prediction. In this work, we are focussed on the last.

\paragraph{Link prediction} is the task of predicting the most likely element given a tuple where one element is missing, e.g., given a triple (\textit{s,p,?}), to predict the most likely $o \in \entitySet$.
With \splime\ we introduce a novel approach that facilitates temporally aware link prediction using static embedding methods. That is, we take the temporal scope of facts into account when performing link prediction. To achieve this, \splime\ transforms every (\textit{s,p,o,h}) tuple into an approximately equivalent ($s,p\text{(}h\text{)},o$) triple, where the specific predicate used is a function of time.


\paragraph{Network proximity measures}\label{sec:proximity} capture the proximity or similarity between a pair of nodes in a graph. In order to do this, two sources of information can be used: the structure of the graph itself or the attributes of the nodes. As KGs do not have node or predicate attributes, we use methods based solely on the structure of the graph. This approach has the advantage that it works for any type of KG. In contrast, an attribute based approach requires a similarity function to be defined for each attribute type.
To be precise, we utilize measures based on \textit{node-neighborhoods}. The node-neighborhood of a node, or entity $s \in \entitySet$, denoted $\Gamma(s)$, is the set of nodes that are direct neighbors of $s$. The idea is that two nodes are more likely to form a link if they have many overlapping neighbors. 
We experiment with three network proximity measures. \textit{Jaccard}, \textit{Adamic/Adar}, and \textit{Preferential attachment}. Their formulations are written as a function $sim(s, o)$ of the entity pair being evaluated as shown below:
\begin{center}
    \begin{tabular}{c|c|c}
	\textbf{Jaccard} & \textbf{Adamic/Adar} & \textbf{Pref. attachment} \\
	\hline
	$sim(s,o)= \frac{|\Gamma(s) \cap \Gamma(o)|}{|\Gamma(s) \cup \Gamma(o)|}$ & 
	$ \sum\limits_{z \in \Gamma(s) \cap \Gamma(o)} \ \frac{1}{\log | \Gamma(z) |} $ & 
	$ | \Gamma(s) | * | \Gamma(o) |$
\end{tabular}
\end{center}
The first measure is the well known Jaccard set similarity metric. The second measure Adamic/Adar was introduced in~\cite{ADAMIC20032113} based on the notion that common features (nodes) should be weighted less heavily than uncommon ones. Lastly, preferential (pref.) attachment argues that the probability that a node obtains a new edge is proportional to its current number of edges. 
In this paper, we use the proximity scores over time as a measure of graph evolution. By considering the proximity scores between entities over time as time series data, we can evaluate when their semantics change significantly. This is done through change point detection.

\paragraph{Change point detection (CPD)} is the problem of finding (abrupt) variations (change points) in time series data \cite{aminikhanghahi2017survey}. Recall that the timestamp set $\timeSet$ contains $l$ elements. CPD considers a time series $S$, i.e., $S = \{x_1,x_2,\dots, x_l\}$, where $x_i$ is a data vector designating the value of the time series at timestamp $i$ with $1 \leq i \leq l$. In our case, we create a time series for every predicate $r$, i.e., we create $S_r \ \forall r \in \predSet$. For a given $S_r$, each entry $x_i$ represents a vector containing similarity scores between nodes that are connected via edges labeled with $r$ at some timestamp $i$.

Truong et al.~\cite{truong2020selective} define CPD as a model selection problem which consists of selecting the best possible segmentation of $S$ such that an associated cost function is minimized. In this definition a CPD algorithm consists of three components: a cost function, a search method and a constraint (on the number of change points). The cost function is a measure of homogeneity, i.e., how similar data points in a given segment are. The search method concerns locating possible segment boundaries, i.e., it locates parts of the signal that should be grouped together. 
When the number of change points is not known a priori a complexity penalty is applied to prevent overfitting to the data. The larger the number of change points, the higher the induced penalty. 
In addition, let $c(\cdot)$ denote a cost function which measures how good a fit is for a sub-signal of the signal $S$. Commonly used cost functions are the $\ell_1$ and $\ell_2$ norms. However, these can only be used when the data exists in a euclidean space. In other cases, one prefers a \textit{kernel function} \cite{garreau2017change}. Since we apply CPD to the output of network proximity measures, whose outputs do not necessarily occupy a euclidean space, \splime\ uses the well-known radial basis kernel function  instead. For a further explanation of CPD and how it relates to \splime , including a full explanation of the cost function used, we refer to Section~\ref{sec:cpdcost} in the Appendix. 

More formally, CPD is defined as finding an optimal segmentation $L = \{k_1, \ldots, k_K\}$, $0 \leq K \leq l \in \mathbb{N}^{+}$ for $S$, i.e., $L$ is a set of indices of elements of $S$ at which change points occur. 
In turn, $\textit{pen}(L)$ measures the complexity associated with the segmentation. 
Let $S_{a,b}$ denote the sub-signal of $S$ between indices $a, b \in \timeSet, a \leq b$. 
The CPD problem for an unknown number of change points can be written as

$~~~~~~~~~~~~~~~~~~~~~~~~~~~~~~\min\limits_L \sum_{k=1}^{K - 1} c(S_{k,k+1}) + \textit{pen}(L). $

In our experiments, we use the bottom-up search function. This procedure starts with many change points and then deletes those that are less significant. This method has two parameters. The first is \textit{minimum size}, which specifies the minimum length of a segment. The second is \textit{jump}, which specifies how many samples apart the initial change points should be distributed.

\section{SpliMe}\label{sec:splime}
We propose \splime\ as a way to include the temporal scope of facts at the level of entities and/or predicates. 
At the core of \splime\ is a transformation function \(f \ : \TKG \mapsto \TKG^\prime \), which takes a TKG and returns a new TKG where temporal scope is (partially) included at the level of entities and/or predicates. 
\(f\) is then applied repeatedly until a stopping criterion is met. 
While \splime\ can be applied to both entities and predicates, preliminary research showed that it works best when applied to the latter. 
Therefore, we will focus on the results on predicates for this paper.
We have developed several different implementations for $f$ categorized into timestamping, splitting and merging approaches. In the coming sections we will provide an in-depth explanation of each approach. Additionally, we provide an algorithm for each method. Due to space constraints, only the algorithm for the CPD-based efficient splitting method is given here, the others are available in the Appendix~\ref{sec:pseudocode}.

\subsection{Timestamping}
Conceptually the simplest approach, timestamping converts each temporal fact in the TKG  into a set of facts, one for each timestamp for which the fact was true. 
A variant of this approach was previously defined in~\cite{derivingvalidity} under the name \textit{Naive-TTransE}. 
Formally, given an injective function \(u(\predSet,\timeSet) \mapsto \predSet^\prime \) which creates a new predicate for every \textit{(predicate, timestamp)} combination, timestamping creates a set of facts for a given quintuple, i.e.
\(f(s,p,o,b,e) = \{(s,u(p,t),o) \ |\ \forall t \in \timeSet \ s.t.\ b \leq t \leq e \} \). 
To obtain a fully timestamped KG, this process must be applied to every fact in a TKG.

\subsection{Splitting}\label{sec:splitting}
Timestamping serves mostly as a baseline method for including the temporal scope of facts to which we can compare our other implementations. The creation of a large number of triples and predicates may cause the resulting KG to become too sparse. As an alternative, we introduce two different approaches: splitting and merging. In this section we firstly discuss the common splitting framework, before displaying our two different splitting approaches. In the next section we will discuss the merging approach.
Given a predicate $r \in \predSet$ and a timestamp $t \in \timeSet$, splitting adds two new predicates, $r_1$ and $r_2$, to the predicate set $\predSet' = \{r_1, r_2\} \cup \predSet$. Each quintuple $(s,p,o,b,e) \in T^{p=r}$ is then updated to contain either $r_1$ if $e \leq t$, or $r_2$ if $b \geq t$. Otherwise, the fact is split into two: $\{(s,r_1,o,b,t), (s,r_2,o,t,e)\}$. As a result, splitting does not necessarily increase the size of the dataset. Instead, splitting reduces the number of facts associated with each predicate.

\paragraph{\textbf{Parameterized splitting}} Formally, parameterized splitting is done via a function $f: (\TKG, c_p, c_t) \mapsto \TKG^\prime$, which given a TKG $\TKG$, splitting criteria $c_p$ which selects a predicate, and criteria $c_t$ which selects a timestamp, maps to a new TKG $\TKG^\prime$. The challenge here lies in defining an effective $c_p$ and $c_t$. For $c_p$, a general approach that produced a good performance in our experimental evaluation is selecting the predicate that occurs the most, i.e., 
$$c_p = \arg\max_{r \in \predSet} | \TKG^{p = r} |. $$ 
Intuitively, this is the predicate that benefits the most by making it more specific.
We propose two methods for $c_t$: \textit{time} and \textit{count}. Let $t_f^r$ and $t_l^r$ denote the first and last timestamps associated with a predicate $r \in \predSet$ in a TKG. The \textit{time method} splits a fact that contains $r$  at $t = (t_f^r + t_l^r) / 2 $. The \textit{count method} selects a point in time that results in the most balanced number of triples on both sides of the split. This is done by counting how many facts end before, or end after a specific timestamp. Formally, the count function can be written as 
$$
c_t(r) = \arg\min_{t \in \timeSet} abs(| \TKG^{p = r, e \leq t} |
-
|T^{p = r, b \geq t} |),
$$ 
where \textit{abs} denotes the absolute value function.
Furthermore, we require that $T^{p = r, e \leq t} \neq \emptyset$ and $ \TKG^{p = r, b \geq t} \neq \emptyset$ in order to prevent the consideration of non-existing (\textit{predicate, timestamp}) combinations. We have performed experiments with both methods and found that both achieve similar results. Which one performs best depends on the dataset used. Regarding stop condition, we continue splitting until the  number of predicates in the KG has grown by a factor $Grow$.

\paragraph{\textbf{CPD-based efficient splitting}}\label{sec:cpdsplit}
The parameterized splitting method described above uses relatively simple measures to decide where, and when, to apply splits. However, there is much more information present in the structure of the KG that could be used to inform the splitting procedure. Thus, we introduce CPD-based efficient splitting: a two-step process to determine how to place split points. 

In the first step, network proximity measures are applied to a TKG to calculate a \textit{signature} vector for every predicate, at every timestamp, i.e., we create a time series $S_r$, $\ \forall r \in \predSet$.
For any $S_r$, each entry represents a vector of proximity scores between pairs of nodes that are connected by $r$ in the graph.
The sequence of entries then represents how the predicate has evolved over time.
The second step is to input these vectors to a CPD algorithm.
Split points are placed at the timestamps where CPD algorithm has identified change points.

To overcome sparsity of information, a TKG is considered as an undirected graph, i.e., we do not take into account whether entities occur as subjects or objects. Signatures are created by slicing the TKG into subgraphs representing the different (\textit{predicate, timestamp}) combinations, i.e., a subgraph $s_{r,t} = \TKG^{p=r, b \leq t \leq e}$ is created for every $(r,t)$ pair.

Let $f_{score}: (\mathcal{E},\mathcal{E}, \TKG) \mapsto \real$ denote a proximity function (e.g. preferential attachment) that calculates the proximity score for a given entity pair on the given subgraph $s_{r,t}$. 
Additionally, let $\textit{pairs}_r$ denote the set of entity pairs for predicate $r$, i.e., $\textit{pairs}_r = \{(e_1,e_2) \ | \ e_1, e_2 \in \entitySet \wedge ( (e_1,p,e_2) \lor (e_2,p,e_1) \in \TKG^{p=r}) \}$.
Now, for every predicate $r$ we calculate the proximity score of each pair in $\textit{pairs}_r$ for every timestamp and add the result to the signature vector.
That is, we calculate $f_{score}(e_1,e_2, s_{r,t}) \ \forall (e_1,e_2) \in pair_r$ and $\ \forall t \in \timeSet$.

Next, we discuss how CPD is applied to the signature vectors. As the number of change points for a given predicate is unknown, a complexity penalty or residual value has to be used to determine the optimal number of split/change points. In our experimental evaluation, we utilize a bottom-up segmentation algorithm in combination with a residual $\epsilon$ in the range [1, 100] depending on the data being transformed. A high $\epsilon$ residual allows for more leeway in reconstructing the data, necessitating less change points. Vice-versa, a low value for $\epsilon$ implies more change points. To ensure that the found change points do not depend on the magnitude of the signatures, we normalize all signatures before passing them into the CPD algorithm. 

\begin{algorithm}[!t] 
    \begin{algorithmic}[1]\caption{: \splime: CPD-based efficient splitting }
        \label{alg:proximity_cpd}
        \Function{cpd\_split}{$\TKG$, $\predSet$, $f_{score}$}
        \Let{\textit{split\_points}}{\{\}}
        \Comment{Stores a list of splits for every predicate.}

        \ForAll{$r \in \predSet$}
        \Let{$\TKG^\prime$}{$\TKG^{p=r}$}
        \Let{\textit{signatures}}{$\textsc{calc\_signatures}(\TKG^\prime, f_{score} )$}
        \Let{$\textit{split\_points}_p$}{$\textit{apply\_cpd}(\textit{signatures})$}
        \EndFor 
        \Let{$\timeSet^{\prime\prime}, \predSet^\prime$}{$\textit{apply\_split}(\TKG,\predSet, \textit{split\_points})$}
        \Comment{\parbox[t]{.44\linewidth}{Each split point can be applied using the logic explained in Section~\ref{sec:splitting}.}} \\

        \Return{$\timeSet^{\prime\prime}, \predSet^\prime$}
        \EndFunction 
        \\
        \Function{calc\_signatures}{$\TKG$, $f_{score}$}
        \Let{\textit{signatures}}{\{\}}
        \Comment{Stores a signature for every timestamp}

        \ForAll{$t \in \timeSet$}
        \Let{$\TKG^\prime$}{$ \TKG^{b \leq t, e \geq t}$}
        \Comment{\parbox[t]{.5\linewidth}{Slice the TKG to contain just the facts that are valid at $t$}}

        \ForAll{$(s,p,o,b,e) \in \TKG^\prime$}
        \Let{\textit{score}}{$f_{score}(s,o,\TKG^\prime) $}
        \Let{$\textit{signatures}_{t_{s,o}}$}{\textit{score}}
        \Comment{Store the score in the signature}
        \EndFor
        \EndFor \\
        \Return{\textit{signatures}}
        \EndFunction
    \end{algorithmic}
\end{algorithm}

Pseudocode for this approach is given in Algorithm~\ref{alg:proximity_cpd}. We have included two functions: one to calculate the signature vectors for a given predicate, and one which uses these signatures to calculate the split points and apply them. The main entry point for the procedure is the \textsc{cpd\_split} function defined on line 1. The \textit{apply\_cpd} and \textit{apply\_split} procedures are not included for legibility.
Firstly, a dictionary or hash set is created which will keep track of all the split points of a given predicate, for every timestamp (on line 2). Between lines 3 and 7 we loop over all predicates, slice the KG (line 4) calculate the signatures using $f_{score}$  (line 5). In the last line of the loop we perform CPD (line 6). Once all the predicates have been processed, we apply the found split points to the TKG and the predicate set (line 8). A new TKG and predicate set are returned on line 9.

Signature creation is done as follows. On line 13 we loop over all timestamps in the KG. In every loop we select only the facts that are valid at the given timestamp (line 14). We then loop over all these facts, calculate the proximity score of the entities it is composed of (line 16) and add that score to the signature vector (line 17). Because we model the evolution of node proximity over time, the location of each entry in the signature vector must be consistent. Therefore, we assume that adding to the signature always places an (s,o) pair at the same index.

\subsection{Merging}
The splitting methods described above refine the temporal scope of the the predicates in the knowledge graph at every iteration by making them more specific. They can therefore be considered top-down approaches. As an alternative we envision a bottom-up approach where, starting with a large selection of possible (\textit{predicate, timestamp}) combinations, refinement is performed by selectively combining pairs. We will now describe such a method called \textit{merging}. 

Merging starts with a timestamped TKG (consisting of $(s,p,o,h)$ quadruples) and selectively merges  predicates which belong to the same original predicate and are subsequent (or contiguous) in time. Given a predicate pair $(r_a, r_b)$ to merge, a new predicate $r_c$ is generated, and every $(s,p,o,h) \in \{ \mathit{T}^{p=r_a} \cup \mathit{T}^{p=r_b} \} $ is then updated to contain $r_c$; lowering the number of unique predicates. By iteratively applying the merging procedure until no more merging candidates exist, one could regenerate the original graph. However, experimental results show that the performance declines when too many predicates are removed.

To select the predicates and timestamps that can be merged, we define two functions. A function $l_p:\predSet^\prime \mapsto \predSet$ which given a predicate of the timestamped TKG returns the source predicate in the original TKG, and a function $l_t:\predSet^\prime \mapsto \timeSet$ which given a predicate of the timestamped TKG returns the time associated with that predicate. A predicate pair $(r_a, r_b)$ is only valid as a merge candidate if i) the source predicates are the same: $l_p(r_a) = l_p(r_b)$, and if ii) there exists no predicate whose timestamp is in between those associated with $r_a$ and $r_b$: $\nexists r_c \in \predSet^\prime \ s.t. \ l_t(r_a) \leq l_t(r_c) \leq l_t(r_b)$. We continue this procedure until the number of predicates in the KG has shrunk by a factor $shrink$ compared to that in the timestamped KG.

\section{Empirical Evaluation}\label{sec:method}
In order to evaluate the effectiveness of \splime\ we have implemented the aforementioned methods and executed them on three TKGs commonly used for the evaluation of TKG embedding methods. In addition to this quantitative analysis, we have performed a qualitative analysis of \splime\  which is available in the Appendix~\ref{sec:qualitative}.
All \splime\ methods were implemented in the Python programming language.
Additionally, we use the Ruptures library~\cite{truong2020selective} for change point detection.
All models were trained using the Ampligraph framework, a suite of machine learning tools used for supervised learning~\cite{ampligraph}.

\subsection{Datasets}\label{sec:datasets}
Experiments were performed on 3 datasets commonly used in TKG embedding literature. An overview of their characteristics is displayed in Table~\ref{tab:dataset_characteristics}. We maintain the original train/test/validation splits for all datasets. 
\paragraph{\textbf{Wikidata12k \& YAGO11k}} are valid-time TKGs used in~\cite{dasgupta2018hyte}. The temporal scopes are set to year-level granularity. For many facts in the data either the beginning or end time is missing. Here, the missing time is set to the first or last timestamp in the TKG respectively. Additionally, there are facts whose temporal scope is invalid either because they cannot be parsed correctly or because the end time is before the start time. These facts are removed. Both of these modifications are default procedure for handling these data sets.
\paragraph{\textbf{ICEWS14}} is an event-based TKG consisting of timestamped political events. It contains all events that occurred in 2014 and was extracted from the complete ICEWS KG by~\cite{garcia-duran-etal-2018-learning}. Because \splime\ operates on quintuples rather than quadruples, ICEWS14 is first converted into the valid-time format by setting $b = e = h$ for every fact.

\begin{table}[t]
    \centering
    \caption{\label{tab:dataset_characteristics}Overview of the characteristics of the used datasets. }
    \begin{tabular}{l||cccccc}
        \textbf{Dataset} & $|\entitySet|$ & $|\predSet|$ & $|\timeSet|$ & $|\textit{train}|$ & $|\textit{valid}|$ & $|\textit{test}|$ \\
        \hline
        YAGO11k          & 10,526         & 10           & 59            & 16,408             & 2,050              & 2,051             \\
        Wikidata12k      & 12,554         & 24           & 70            & 32,497             & 4,062              & 4,062             \\
        ICEWS14          & 7,128          & 230          & 365          & 72,826             & 8,941              & 8,963
    \end{tabular}
\end{table}

\subsection{Baselines}\label{sec:baseline}
We compare the results of our methods with two baselines. The first is the \textit{Vanilla} baseline as discussed in Section~\ref{sec:currentproblems}. This method strips out all temporal scopes from the TKG, turning it into a static KG. The second is the \textit{Random} baseline which works by applying splits randomly across predicates and timestamps in a uniform manner. Specifically, at every split step, a predicate is chosen from the currently existing predicates. Then, a timestamp between that predicates first and last occurrence in the dataset is chosen. This is repeated until the desired number of predicates have been created. For the valid time TKGs we report the average of 7 runs. For ICEWS14 the results are averaged over 5 runs. 
Additionally, we compare \splime\ to the current state of the art: ATiSE~\cite{xu2019temporal}, HyTE~\cite{dasgupta2018hyte}, TTransE~\cite{jiang2016encoding}, and TDG2E~\cite{Tang2020TimespanAwareDK} for the valid-time TKGs, and ATiSE, DE-SimplE~\cite{goel2020diachronic}, HyTE, TA-DistMult~\cite{garcia-duran-etal-2018-learning} and TNT-ComplEx~\cite{lacroix2020tensor} for the event-based TKGs. 

\subsection{Hyperparameters}
While \splime\ is model-agnostic and can be combined with any KG embedding method, we use TransE in our experiments and compare with TKG embedding models that are mostly built on top of the TransE family. 
Our model hyperparameters are adapted from \cite{garcia-duran-etal-2018-learning}. We use the same model hyperparameters for all experiments. Specifically, we train for 200 epochs with embedding size = 100 and learning rate = $10^{-3}$. We set the batch size to 500 and generate 500 negative samples per batch. Optimization was done using the ADAM optimizer in combination with a self-adversarial loss function \cite{kingma2014adam}. For CPD-based splitting, we use a kernelized mean change cost function with bottom-up search where jump size and minimal section length are both set to 1.

Because \splime~ applies transformations to TKGs, we also have data hyperparameters. For both the \textit{time} and \textit{count} splitting methods we experimented with $Grow \in \{ 5, 10, 15, 20, 25, 30\}$ for all data sets. Regarding merging, we tested $Shrink \in \{ 1.5, 2, 4, 6, 8, 10\}$. Lastly, \textit{CPD-based splitting} was evaluated using the Jaccard, Adamic/Adar (\textit{Adar}) and Preferential  attachment (\textit{Pref}) proximity measures, with $\epsilon$ (residuals) $\in \{1.25, 2.5, 5, 10, 15, 20\}$ for Wikidata12k and YAGO11k, and $\epsilon \in \{5, 12.5, 25,50,100,150,200\}$ for ICEWS14.

\subsection{Evaluation Procedure}
KG embeddings are evaluated on the link prediction task as described in \cite{bordes2013translating}. For a triple in the test set (\textit{s,p,o}) either the subject or object is replaced with all $e \in \entitySet$. That is, subject and object evaluation is combined. In line with the \textit{filtered} setting, we remove any resulting triples which occur in the train, test or validation sets. All created triples are then scored by the model and the result is sorted. The rank of the original triple is recorded. This process is repeated for all triples, giving us a set of ranks $R$.
From this set we calculate two metrics. These are mean reciprocal rank (MRR) and hits@k for $k \in \{1,3,10\}$. These are defined as:
\begin{center}
\begin{tabular}{c|c}
	\textbf{MRR} & \textbf{Hits@k} \\
	\hline
	$\frac{1}{|R|} \sum_{h \in R}\frac{1}{h}$ $\hphantom{sp}$  & $\hphantom{sp}$ $\frac{1}{|R|}\sum_{h \in R} \begin{cases}
		1 \  \text{if} \ h \leq k \\
		0 \  \text{otherwise} 
	\end{cases} $
\end{tabular}
\end{center}

Finally, all results are obtained using the \textit{inter} filter setting as described in Section~\ref{sec:currentproblems}.

\subsection{Results}\label{sec:result}
\begin{table}[!t]
    \caption{Overview of the optimal result obtained for each method. The best results are highlighted in \textbf{bold}.}\label{tab:main_result}
	\begin{subtable}{\textwidth}
		\caption{Wikidata12k }
		\centering
		\begin{tabular}{ll||cccc|c}
			\textbf{Method} & \textbf{Setting}                & \textbf{MRR}   & \textbf{Hits@1} & \textbf{Hits@3} & \textbf{Hits@10} & \textbf{\# Preds} \\
			\hline
			Vanilla         & -                               & 0.209          & 12.4\%          & 22.7\%          & 37.9\%           & 24                \\
			Random          & -                               & 0.289          & 17.9\%          & 33.3\%          & 50.7\%           & 423               \\
			\hline
			Timestamp       & -                               & 0.340          & 21.1\%          & 40.8\%          & 58.1\%           & 1622              \\
			Split (time)    & $\textit{Grow}=10$              & 0.320          & 20.1\%          & 37.2\%          & 54.9\%           & 240               \\
			Split (count)   & $\textit{Grow}=25$              & 0.300          & 18.3\%          & 34.5\%          & 53.8\%           & 600               \\
			Split (CPD)       & \textit{Pref}, $\epsilon = 2.5$ & 0.328          & 20.9\%          & 38.1\%          & 56.0\%           & 726			\\
			Merge           & $\textit{Shrink}=4$             & \textbf{0.358} & \textbf{22.2\%} & \textbf{43.3\%} & \textbf{61.0\%}  & 423               
		\end{tabular}
	\end{subtable}
	\begin{subtable}{\textwidth}
		\caption{YAGO11k }
		\centering
		\begin{tabular}{ll||cccc|c}
			\textbf{Method} & \textbf{Setting}              & \textbf{MRR}   & \textbf{Hits@1} & \textbf{Hits@3} & \textbf{Hits@10} & \textbf{\# Preds} \\
			\hline
			Vanilla         & -                             & 0.188          & 8.2\%           & 23.8\%          & 35.6\%           & 10                \\
			Random          & -                             & 0.197          & 7.8\%           & 25.2\%          & 39.9\%           & 200               \\
			\hline
			Timestamp       & -                             & 0.197          & 6.9\%           & 26.0\%          & 41.2\%           & 570               \\
			Split (time)    & $\textit{Grow}=20$            & 0.213          & \textbf{9.0\%}  & 27.0\%          & 43.2\%           & 200               \\
			Split (count)   & $\textit{Grow}=25$            & 0.196          & 8.1\%           & 24.1\%          & 40.3\%           & 250               \\
			Split (CPD)       & \textit{Pref}, $\epsilon = 5$ & \textbf{0.214} & 6.5\%           & \textbf{29.9\%} & \textbf{45.8\%}  & 177			\\
			Merge           & $\textit{Shrink}=2$           & 0.195          & 6.2\%           & 26.3\%          & 42.0\%           & 290             
		\end{tabular}
	\end{subtable}
	\begin{subtable}{\textwidth}
		\caption{ICEWS14 }
		\centering
		\begin{tabular}{ll||cccc|c}
			\textbf{Method} & \textbf{Setting}               & \textbf{MRR}   & \textbf{Hits@1} & \textbf{Hits@3} & \textbf{Hits@10} & \textbf{\# Preds} \\
			\hline
			Vanilla         & -                              & 0.141          & 0.1\%           & 18.9\%          & 42.2\%           & 230               \\
			Random          & -                              & 0.172          & 2.0\%           & 23.4\%          & 48.1\%           & 3500              \\
			\hline
			Timestamp       & -                              & \textbf{0.213} & \textbf{4.7\%}  & \textbf{29.4\%} & \textbf{54.4\%}  & 17061             \\
			Split (time)    & $\textit{Grow}=20$             & 0.190          & 3.0\%           & 26.3\%          & 51.6\%           & 4600              \\
			Split (count)   & $\textit{Grow}=25$             & 0.191          & 3.0\%           & 26.2\%          & 52.2\%           & 5750              \\
			Split (CPD)       & \textit{Adar}, $\epsilon = 25$ & 0.196          & 2.9\%           & 27.3\%          & 53.0\%           & 5866			\\
			Merge           & $\textit{Shrink}=1.5$          & 0.207          & 4.1\%           & 28.7\%          & 53.9\%           & 11449             
		\end{tabular}
	\end{subtable}
\end{table}

The best results displayed in Table~\ref{tab:main_result} show that all methods, even the random baseline, provide an increase in performance on all metrics compared to the vanilla baseline. Therefore, we can say that incorporating time at the level of predicates improves the link prediction capabilities of static KGE models. Furthermore, all \splime\ methods have improved performance compared to the random baseline, suggesting that they indeed capture the temporal scope of facts in a more efficient manner.

On Wikidata12k the best result is achieved using the merge method creating a version of the dataset with 423 predicates (17.6 times increase). Here, merging outperforms all other approaches on every recorded metric. It outperforms Vanilla/TransE and the random baseline by approximately 23\% and 10\%  on the hits@10  respectively. This represents a 60\% and 20\% increase in performance respectively.

On YAGO11k the best result is achieved using CPD in combination with preferential attachment as a proximity function. Notably, this method produces 177 predicates (17.7 times increase), which are 113 and 23 fewer than the best merge and split approaches respectively. \textit{CPD-based splitting} outperforms the vanilla baseline by more than 10\%  on the hits@10, which is a 28\% increase in performance.

Lastly, on ICEWS14 the best result is achieved with the timestamping approach, which achieves the highest scores on all metrics. Timestamping outperforms the random baseline by 6\% (13\% increase) on the hits@10. On the hits@1 metric, the difference is 2.7\%, representing a 135\% increase. Compared to our vanilla TransE baseline the results are better, especially on hits@1 as the vanilla baseline scores just 0.1\%.

\begin{table}[!t]
    \caption{\label{tab:result_comparison}Comparison of the best \splime\ approach, for each dataset, and the current state of the art. Best results are highlighted in \textbf{bold}. The method column contains the model name along with the paper that result was reported in.}
	\begin{subtable}{\textwidth}
        \caption{On Wikidata12k, the best \splime\ result was achieved with the \textit{merge} method. On YAGO11k, the best \splime\ result was achieved using \textit{CPD-based splitting}.}
		\centering
		\begin{tabular}{l||cccc||cccc}
             & \multicolumn{4}{c}{Wikidata12k} &  \multicolumn{3}{c}{YAGO11k} \\
			\textbf{Method} & \textbf{MRR}   & \textbf{Hits@1} & \textbf{Hits@3} & \textbf{Hits@10} & \textbf{MRR}   & \textbf{Hits@1} & \textbf{Hits@3} & \textbf{Hits@10} \\
			\hline
			\splime         & \textbf{0.358} & \textbf{22.2\%} & \textbf{43.3\%} & \textbf{61.0\%}  & \textbf{0.214} & 6.5\%           & \textbf{29.9\%} & \textbf{45.8\%}  \\
			\hline
			ATiSE~\cite{xu2019temporal}       & 0.252          & 14.8\%          & 28.8\%          & 46.2\%           & 0.185          & \textbf{12.6\%} & 18.9\%          & 30.1\%           \\
			HyTE~\cite{xu2019temporal}        & 0.180          & 9.8\%           & 19.7\%          & 33.3\%           & 0.105          & 1.5\%           & 14.3\%          & 27.2\%           \\
			TTransE~\cite{xu2019temporal}     & 0.172          & 9.6\%           & 18.4\%          & 32.9\%           & 0.108          & 2.0\%           & 15.0\%          & 25.1\%           \\
			TDG2E~\cite{Tang2020TimespanAwareDK}           & -              &       -          & -               & 40.2\%           & -              & -               & -               & 31.1\%
		\end{tabular}
	\end{subtable}
	\begin{subtable}{\textwidth}
		\centering
        \caption{On ICEWS14, the best  \splime\ result was achieved with the \textit{timestamp} method.}
		\begin{tabular}{l||cccc}
			\textbf{Method} & \textbf{MRR}   & \textbf{Hits@1} & \textbf{Hits@3} & \textbf{Hits@10} \\
			\hline
			\splime         & 0.213          & 4.7\%           & 29.4\%          & 54.4\%           \\
			\hline
			ATiSE~\cite{xu2019temporal}       & 0.545 & 42.3\% & \textbf{63.2\%} & \textbf{75.7\%}  \\
            DE-SimplE~\cite{goel2020diachronic}       & 0.526          & 41.8\%          & 59.2\%          & 72.5\%           \\
			HyTE~\cite{xu2019temporal}        & 0.297          & 10.8\%          & 41.6\%          & 60.1\%           \\
			TA-DistMult~\cite{garcia-duran-etal-2018-learning}     & 0.477          & 36.3\%          & -               & 68.6\%			\\
			TNTComplEx~\cite{lacroix2020tensor}		& \textbf{0.56}	& \textbf{46\%} &  61\% & 74\%
		\end{tabular}
	\end{subtable}
\end{table}

In Table~\ref{tab:result_comparison} we compare the best \splime\ results with the current state-of-the-art models. The results show that \splime\ outperforms these models on the valid time datasets. In fact, \splime\ performs 15 percentage points better than the second-best model (ATiSE) on the hits@10 metric for both Wikidata12k and YAGO11k. Only on the hits@1 metric on the YAGO11k dataset it is outperformed by ATiSE. Yet, it still outperforms the other models. The ICEWS14 results are not as good. Here, \splime\ is outperformed by the other models. However, we still observe a noteworthy increase in performance compared to our vanilla baseline. This shows that \splime\ works best with valid time TKG, where it is possible to split and merge the temporal facts by leveraging their temporal scopes.

\section{Test Leakage in Current Literature}\label{sec:currentproblems}
We found that the results reported in the state of the art did not remove duplicates when stripping temporal scopes from KG facts~\cite{goel2020diachronic,jin2020recurrent}. To elaborate, when temporal KGs are embedded using static methods, the temporal scope is simply discarded. However, we note that this leads to duplicate information which distorts the evaluation result. To give an example, consider the following two facts: (\textit{Obama, visited, the Netherlands, 2014}) and (\textit{Obama, visited, the Netherlands, 2018}). When temporal scope is stripped these both result in (\textit{Obama, visited, the Netherlands}).
This causes two issues. Firstly, there may exist duplicate triples inside any given train/test/test split (\textit{intra-set}). In the train split this causes the model to fit to multiple instances of the same triple. In the test split the model's performance will be abnormally determined by such triples. Secondly, examples may be duplicated between splits (\textit{inter-set}), causing test-leakage.

\begin{table}[t]
	    \centering
    \caption{\label{tab:dataset_dupes}Overview of the number of inter and intra set duplicates after stripping temporal scopes for each of the used datasets.  YAGO is unaffected.}
    \begin{tabular}{l||ccc}
                                       & \textbf{Wikidata12k} & \textbf{YAGO11k} & \textbf{ICEWS14} \\
        \hline
        \# duplicates in train (\%)    & 4720 (14.53\%)       & 0 (0\%)          & 30136 (41.38\%)  \\
        \# duplicates in test (\%)     & 214 (5.27\%)         & 0 (0\%)          & 1544 (17.23\%)   \\
        \# duplicates in valid (\%)    & 193 (4.75\%)         & 0 (0\%)          & 1610 (18.01\%)   \\
        \hline
        \# test triples in train (\%)  & 1042 (27.10\%)       & 0 (0\%)          & 3499 (47.16\%)   \\
        \# valid triples in train (\%) & 1027 (26.54\%)       & 0 (0\%)          & 3527 (48.11\%)
    \end{tabular}
\end{table}

Table~\ref{tab:dataset_dupes} contains an overview of the number of duplicate triples in 3 datasets commonly used for evaluating TKG embedding models. Our analysis shows that YAGO11k is unaffected by this duplication issue. However, Wikidata12k and ICEWS14 are affected, with the latter having almost half of its test triples appear in the training set. We hypothesize that this is due to its comparatively large number of timestamps.
To counteract this issue we introduce two new filtering methods. \textit{Intra-set filtering} which filters out any duplicate triples inside a given split, and \textit{inter-set filtering} which removes any triples from the test/validation split if they occur in the train set. Additionally, the \textit{both} option applies both filtering methods. We investigate the effect of these measures on the performance of the TransE model in Table~\ref{tab:filtered_test}. 

Specifically, we have evaluated the performance of TransE on 2 datasets using 3 different implementations: HyTE~\cite{dasgupta2018hyte}, RotatE~\cite{sun2018rotate}, and Ampligraph~\cite{ampligraph}. Hyperparameters for the different implementations are given in the Appendix~\ref{sec:hyperparm}.
\begin{table}[t]
    \centering
    \caption{\label{tab:filtered_test}Results on 3 different TransE implementations, combined with 4 different filter settings, on both affected datasets.} 
    \begin{tabular}{ll||cccc}
                            &                         & \multicolumn{4}{c}{\textbf{Hits@10 performance}}                                                   \\
        \textbf{Dataset} & \textbf{Implementation} & \textbf{No Filter}                               & \textbf{Inter} & \textbf{Intra} & \textbf{Both} \\
        \hline
        Wikidata12k      & HyTE (TransE)                    & 10.1\%                                           & 3.8\%          & 5.3\%          & 3.5\%         \\
                            & RotatE (TransE)              & 53.8\%                                           & 36.4\%         & 52.4\%         & 33.9\%        \\
                            & Ampligraph        (TransE)      & 52.7\%                                           & 37.7\%         & 52.8\%         & 37.9\%        \\
\hline
        ICEWS14          & HyTE (TransE)                    & 54.1\%                                           & 25.7\%         & 40.6\%         & 23.7\%        \\
                            & RotatE (TransE)                  & 70.3\%                                           & 38.4\%         & 65.1\%         & 37.9\%        \\
                            & Ampligraph     (TransE)         & 56.1\%                                           & 42.2\%         & 56.1\%         & 42.2\%
    \end{tabular}
\end{table}
On Wikidata12k and ICEWS14 the performance on the link prediction task drops significantly. This effect occurs regardless of the implementation used, although the size of the effect does differ between the implementations. On ICEWS14 the performance of the HyTE and RotatE TransE implementations almost halve between no filter and both filtering methods applied, whereas Ampligraph reaches 75\% of the original result. On Wikidata12k the effect is slightly less due to the fact that it has fewer duplicate triples.

\section{Related Work}\label{sec:relatedwork}
In this section, we will provide a brief overview of related work in the area of TKG embedding. For an overview of static embedding methods we refer to the Appendix~\ref{sec:extendedrelatedwork} and/or survey~\cite{ji2020survey}.
TA-TransE~\cite{garcia-duran-etal-2018-learning} uses a recurrent neural network (LSTM) to learn time-aware representations of predicates. (Temporal) predicates are strings, concatenated with sequence of temporal tokens. Valid time is modelled as two predicates, each combined with either the token `occursSince' or `occursUntil'. Like \splime\, TA-TransE is model-agnostic. However, where \splime\ can be said to use an adaptive granularity, TA-TransE represents facts at their original granularity.
HyTE~\cite{dasgupta2018hyte} is an extension of TransH. It considers a TKG as a series of static KG snapshots, where each snapshot represents a point in time containing only the set of facts that were valid at that time. Instead of projecting embeddings on a predicate specific hyperplane, it projects them on a timestamp-specific hyperplane. Unlike \splime\, the number of static KG snapshots is fixed over all predicates.
ATiSE~\cite{xu2019temporal} represents entities and predicates as a multi-dimensional Gaussian distribution. An embedding at a specific timestamp is represented as the mean of the distribution and its uncertainty is represented as the covariance of the distribution. Evolution of entities and predicates is modeled through additive time series decomposition.
Diachronic Embeddings~\cite{goel2020diachronic} include temporal scopes directly in the embedding vector by applying a time-dependent function to a subset of each vector. In effect, the first $k$ parameters in the vector are reserved as representing temporal scopes, where $k$ is a hyperparameter. However, their model can only be applied to event-based KGs, whereas \splime\ can be applied to both event-based and valid-time knowledge graphs.
TDG2E~\cite{Tang2020TimespanAwareDK} utilizes a Gated Recurrent Unit (GRU) to model the dependencies between subsequent temporal slices of the TKG. This gate also models the length of the timespan between two subgraphs. Specifically, this timespan aware temporal evolution model is added to the HyTE model, resulting in a model that can encode temporal scopes and the evolution of that scope over time.
TNTComplEx~\cite{lacroix2020tensor} is the extension of ComplEx~\cite{pmlr-v48-trouillon16} to temporal KGE. TNTComplEx considers a TKG as a 4-order tensor, i.e., a new dimension is added to the tensor where every entry represents a timestamp. Embeddings are learned through tensor decomposition.

\section{Conclusion}\label{sec:conclusion}
In this paper we have introduced \splime, a model-agnostic method that uses static KGE models to embed TKGs. \splime\ operates through selectively splitting and merging predicates. We have shown that even incorporating temporal scope by randomly applying splits improves the link prediction performance of TransE, indicating the viability of the inclusion of temporal scopes and the power of our approach.

All our methods achieve results above  the vanilla TransE and  random baselines. Our experiments show that \splime\ achieves state-of-the-art results in the link prediction task on two datasets commonly used for TKG embedding model evaluation (Wikidata12 and YAGO11k). In addition, it  outperforms our baselines on the ICEWS14 dataset. We have further shown the strength of our results through various qualitative analyses. 
Additionally, we introduced two explicit filtering methods to deal with the issue we uncovered in the procedure used for evaluating static KGE models on TKGs.
As a future work, we focus on two directions: (i) we carry out experiments on larger subsets of Wikidata and YAGO datasets as well as (ii) we use one of the best performing static KGE models (e.g. ComplEx) to see if our results can be improved. 

\bibliographystyle{splncs04}
\bibliography{splime}

\appendix
\section{Framework Hyperparameters}\label{sec:hyperparm}
In the following table, we provide an overview of the applied hyperparameters used for the three different models. Note that RotatE uses steps instead of epochs as a stopping condition.
\begin{table}[H]
    \centering
       \begin{tabular}{l||c|c|c}
        \textbf{Parameter}   & \textbf{HyTE} & \textbf{RotatE}  & \textbf{Ampligraph} \\
        \hline
        Dimension            & 100           & 100              & 100                 \\
        Margin               & 1             & 24               & 1                   \\
        Neg samples          & 5             & 500              & 500                 \\
        Norm                 & $l_2$         & $l_1$            & $l_1$               \\
        Batch size           & 50,000        & 512              & ~500                \\
        Learning rate        & $10^{-4}$     & $10^{-3}$        & $10^{-3}$           \\
        Loss                 & Pairwise      & Self-adversarial & Self-adversarial    \\
        Sampling temperature & -             & 1.0              & 0.5                 \\
        Initializer          & Xavier        & Uniform          & Xavier              \\
        Steps/Epochs         & 500           & 150k             & 200
    \end{tabular}
\end{table}

\section{Algorithms}\label{sec:pseudocode}
In this section, we present the algorithms for the three methods (timestamping, parameterized splitting and merging) that are used by \splime\ for TKG transformation. 
\begin{algorithm}[h]
    \begin{algorithmic}[1]\caption{: \splime: timestamping\ }
        \label{alg:timestamping}
        \Function{Timestamp}{$\TKG$, $\predSet$ $u$}

        \Let{$\TKG^\prime$}{$\{ \}$}
        \Let{$\predSet^\prime$}{$\{ \}$}

        \ForAll{$(s,p,o,b,e) \in \TKG$}
        \For{$t$ = $b$; $t \leq e$ ; $t\texttt{++}$}
        \Let{r}{$u(p, t)$}
        \Comment{\parbox[t]{.5\linewidth}{Lookup/create new a predicate for this (predicate, timestamp) combination}}
        \Let{$\predSet^\prime$}{$\predSet^\prime \cup r $}
        \Let{$\TKG^\prime$}{$\TKG^\prime \cup (s,r,o, t)$}
        \EndFor
        \EndFor \\
        \Return $\TKG^\prime, \predSet^\prime$
        \EndFunction

    \end{algorithmic}
\end{algorithm}

The pseudocode for the timestamping algorithm is given in Algorithm~\ref{alg:timestamping}. We start off by initializing an empty predicate set and temporal knowledge graph in lines 2 and 3. Then, for every quintuple in the knowledge graph, we iterate over all all timestamps in its valid time (line 4). For each such timestamp we create and add a new predicate (to the predicate set), and then add the modified fact to the TKG (lines 7, 8).

\begin{algorithm}[tbph]
    \begin{algorithmic}[1]\caption{: \splime: Parameterized splitting\ }
        \label{alg:splitting}
        \Function{Split}{$\TKG$, $\predSet$, $c_t$} 

        \While{\textbf{not} \textit{stop condition met}}

        \Let{r}{$\arg\max_{r \in \predSet} |T^{p = r}|$}
        \Let{t}{$c_t(|T^{p = r}|)$}
        \Comment{Using either \textit{Time} or \textit{Count} method}
        \Let{$\predSet$}{$\predSet \cup \{ r_1, r_2 \}$}

        \ForAll{$ (s,p,o,b,e) \in \TKG^{p=r}$}
        \Let{$\TKG$}{$\TKG \setminus \{ (s,p,o,b,e) \} $}
        \Comment{Remove the original fact from the TKG}
        \If{$b \leq t \leq e$}
        \Comment{If fact spans the split time}

        \Let{$\TKG$}{$\TKG \cup (s,r_1,o,b,t)$}
        \Let{$\TKG$}{$\TKG \cup (s,r_2,o,t,e)$}
        \ElsIf{$ e \leq t $}
        \Comment{Fact ends before split}
        \Let{$\TKG$}{$\TKG \cup (s,r_1,o,b,e)$}
        \Else
        \Comment{Fact begins after split}
        \Let{$\TKG$}{$\TKG \cup (s,r_2,o,b,e)$}
        \EndIf

        \EndFor
        \Let{$\predSet$}{$\predSet \setminus \{ r \}$ }
        \Comment{Remove the original predicate}

        \EndWhile \\
        \Return $\TKG, \predSet$
        \EndFunction
    \end{algorithmic}
\end{algorithm}

Pseudocode for the splitting approach is given in Algorithm~\ref{alg:splitting}. Continuing until the stop condition is met, each iteration of the algorithm starts by finding the most common predicate in the TKG (line 3) and selecting a timestamp based on the either the time or count method (line 4). On line 5, two new predicates are created: the first represent the predicate until the split timestamp $t$, the second represents it from that point forward. Then for every fact in the TKG which contains the original predicate, we remove that fact in favour of the new 'split' facts. Which predicate ($r_1$, $r_2$) is used depends on the valid time of the quintuple. Finally, we remove the original predicate from the predicate set (line 17).

\begin{algorithm}[tbph]
    \begin{algorithmic}[1]\caption{: \splime: merging \ }
        \label{alg:merging}
        \Function{Merge}{$\TKG$, $\predSet$} 

        \Let{$\TKG^\prime, \predSet^\prime$}{$\textsc{timestamp(}\TKG\textsc{)}$}
        \Comment{Apply timestamping procedure in Algorithm 1}

        \While{\textbf{not} \textit{stop condition met}}
        \Let{$\mathcal{O}$}{\textit{all possible merge options}}
        \Let{$(r_1, r_2)$}{$\arg\min_{(r_1, r_2) \in \mathcal{O}} | \TKG^{\prime \ p=r_1} \cup \TKG^{\prime \ p=r_2} | $}
        \Comment{\parbox[t]{.3\linewidth}{Select the least occuring pair}}
        \Let{$r_n$}{\textit{generate new predicate}}

        \ForAll{$(s,p,o,h) \in \TKG^{\prime \ p=r_1} \cup \TKG^{\prime \ p=r_2}$}
        \Let{$\TKG^\prime$}{$\TKG^\prime \setminus \{(s,p,o,h)\} $}
        \Let{$\TKG^\prime$}{$\TKG^\prime \cup \{(s,r_n,o,h)\}$}
        \Comment{Replace all occurrences of the predicates}
        \EndFor

        \Let{$\predSet^\prime$}{$\predSet^\prime \cup \{ r_n \}$}
        \Comment{Add the new predicate}
        \Let{$\predSet^\prime$}{$\predSet^\prime \setminus \{r_1, r_2 \}$}
        \Comment{Remove the old predicates}

        \EndWhile \\
        \Return{$\TKG^\prime, \predSet^\prime$}

        \EndFunction
    \end{algorithmic}
\end{algorithm}

Pseudocode for the merging approach is given in Algorithm~\ref{alg:merging}. The first step is to apply the \splime\ timestamping method (line 2). The following procedure is performed until the stop condition is met. On line 4 we generate all possible merge options according to the constraints laid out above. From this, we select the pair that occurs the least (line 6). Next, we find all facts in the TKG containing either predicate selected to merge, and iterate over them on line 9. Each of these we remove from the TKG (line 8) and then re-add it (line 9) with the predicate replaced for the merged predicate we created on line 5. Once we have iterated over all predicates we exit the loop. Then we update the predicate set by adding the new, and removing the old predicate (lines 11, 12)

\section{Qualitative Analysis}\label{sec:qualitative}
To strengthen the claim that \splime\ improves embeddings, we perform a qualitative analysis of the result. We investigate this through two avenues. The first is link prediction. Specifically we perform predicate prediction and compare the results from a vanilla TransE model and a \splime\ TransE model. The second avenue is through a t-SNE plot, which allows visualization of the embeddings learned by a model. Each method will be explained in further detail in the relevant section.

\subsection{Link Prediction}
In this section we perform qualitative analysis with regards to the predicate prediction task. (Temporal) predicate prediction is defined analogously to entity prediction. Instead of replacing the subject and object with every entity, the predicate is replaced with every other predicate. That is, given an (\textit{s,?,o,b,e}) quintuple we are tasked with predicting the most likely predicate. Since \splime\  operates on triples, we cannot directly pass (\textit{s,p,o,b,e}) quintuples or (\textit{s,p,o,h}) quadruples to the model. However, we can include the temporal aspect by filtering any answers which are not of the correct temporal scope. Specifically, we feed the model with a head and tail entity and ask for the 25 most likely predicates. From this list of predicates we then remove any predicate whose time span does not at least partially overlap with the time span of the original quintuple.

A selection of such queries is shown in Table~\ref{tab:qualitative_prediction}. Here, the top two results are shown for every query. The correct answer is highlighted in bold. These queries were performed on a vanilla dataset and on a dataset transformed with the split (time) method with growth set to 20. The first four questions are the same as in the original HyTE paper. From these results, it appears that the vanilla TransE model and \splime\ seems to achieve equal results. However, we do note that our TransE vanilla model also performs better than the one used in the original HyTE paper.

Examples where \splime\ works best are those where similar predicates are queried, but at different timestamps. For example, when a person has both the `wasBornIn` and `diedIn` predicates, but at (significantly) different timestamps. Unfortunately, the YAGO11k test set contains just one such example. Instead, the few people for which both the `wasBornIn` and `diedIn` relation are present in the test set have them occur in (almost) the same internal timestamp. Therefore, we believe that this qualitative analysis does not paint the full picture, and \splime\ will outperform TransE in a more exhaustive test.

\begin{table}[htbp]
    \begin{adjustbox}{addcode={\begin{minipage}{\width}}{\caption{A comparison between link prediction results generated by a vanilla TransE model, and the best \splime\ model on the YAGO11k dataset. Some entities have had their names shortened for readability. Queries above the horizontal line equal to those in \cite{dasgupta2018hyte}. Below the horizontal line are original. All examples were taken from the test set to ensure that the model has not seen them before.\label{tab:qualitative_prediction}}\end{minipage}},rotate=90,center}
        \centering
        \begin{tabular}{l||l|l}
            \textbf{Original quintuple (s,?,o,b,e)}                            & \textbf{TransE}                      & \textbf{SpliMe}                      \\
            \hline
            \hline
            G. Carroll, \textit{wasBornIn}, Baltimore, 1928, 1928              & \textbf{wasBornIn}, DiedIn           & \textbf{wasBornIn}, diedIn           \\
            S.A. Laubenthal, \textit{diedIn}, Washington., 2002, 2002          & \textbf{diedIn}, BornIn              & \textbf{diedIn}, isMarriedTo         \\
            E. G. Sander, \textit{graduatedFrom}, Cornell Univ., 1959, 1965    & worksAt, \textbf{graduatedFrom}      & worksAt, \textbf{graduatedFrom}      \\
            E. Maceda, \textit{isAffiliatedTo}, Nacionalista Party, 1971, 1987 & \textbf{isAffiliatedTo}, isMarriedTo & \textbf{isAffiliatedTo}, isMarriedTo \\
            \hline
            A.Rothschild, \textit{isMarriedTo}, E.J. Rothschild, 1877, 2020    & \textbf{isMarriedTo}, wasBornIn      & \textbf{isMarriedTo}, owns           \\
            A.Rothschild, \textit{diedIn}, Paris, 1935, 1935                   & \textbf{diedIn}, wasBornIn           & \textbf{diedIn}, wasBornIn           \\
            E.J. Rothschild, \textit{wasBornIn}, Boulogne-Bi., 1845 , 1845     & \textbf{wasBornIn}, diedIn           & \textbf{wasBornIn}, diedIn           \\
            E.J. Rothschild, \textit{diedIn}, Boulogne-Bi., 1934, 1934         & wasBornIn, \textbf{diedIn}           & \textbf{diedIn}, wasBornIn           \\
            Vilhelm Aubert, \textit{worksAt}, University of Oslo, 1954, 1988   & \textbf{worksAt}, graduatedFrom      & \textbf{worksAt}, graduatedFrom      \\
            Marie Curie, \textit{isMarriedTo}, Pierre Curie, 1859, 1906        & \textbf{isMarriedTo}, created        & \textbf{isMarriedTo}, created        \\
            Marie Curie, \textit{hasWonPrize}, W.G. Award, 1921, 1921          & \textbf{hasWonPrize}, wasBornIn      & \textbf{hasWonPrize}, isMarriedTo    \\
            Marie Curie, \textit{wasBornIn}, Warsaw, 1867, 1867                & \textbf{wasBornIn}, diedIn           & \textbf{wasBornIn}, diedIn           \\
            Pierre Curie, \textit{wasBornIn}, Paris, 1859, 1859                & diedIn, \textbf{wasBornIn}           & \textbf{wasBornIn}, diedIn           \\
            Pierre Curie, \textit{diedIn}, Paris, 1906, 1906                   & \textbf{diedIn}, wasBornIn           & wasBornIn, \textbf{diedIn}           \\
            King-Sun Fu, \textit{worksAt}, MIT, 1961, 2020                     & \textbf{worksAt}, graduatedFrom      & \textbf{worksAt}, graduatedFrom      \\
            King-Sun Fu, \textit{graduatedFrom}, Univ. of Toronto, 1955, 2020  & worksAt, \textbf{graduatedFrom}      & worksAt, \textbf{graduatedFrom}
        \end{tabular}
    \end{adjustbox}
\end{table}

\subsection{t-SNE plots}\label{sec:t_sne_plots}
t-distributed stochastic neighborhood embedding (t-SNE) is a non-parametric high-dimensional data visualization technique.
Each high-dimensional data point is mapped to a location on a two or three dimensional map \cite{JMLR:v9:vandermaaten08a}.
We will use t-SNE to provide a legible visualization of the high-dimensional predicate embeddings learned by our model.

\begin{figure}
    \centering
    \includegraphics[width=\textwidth]{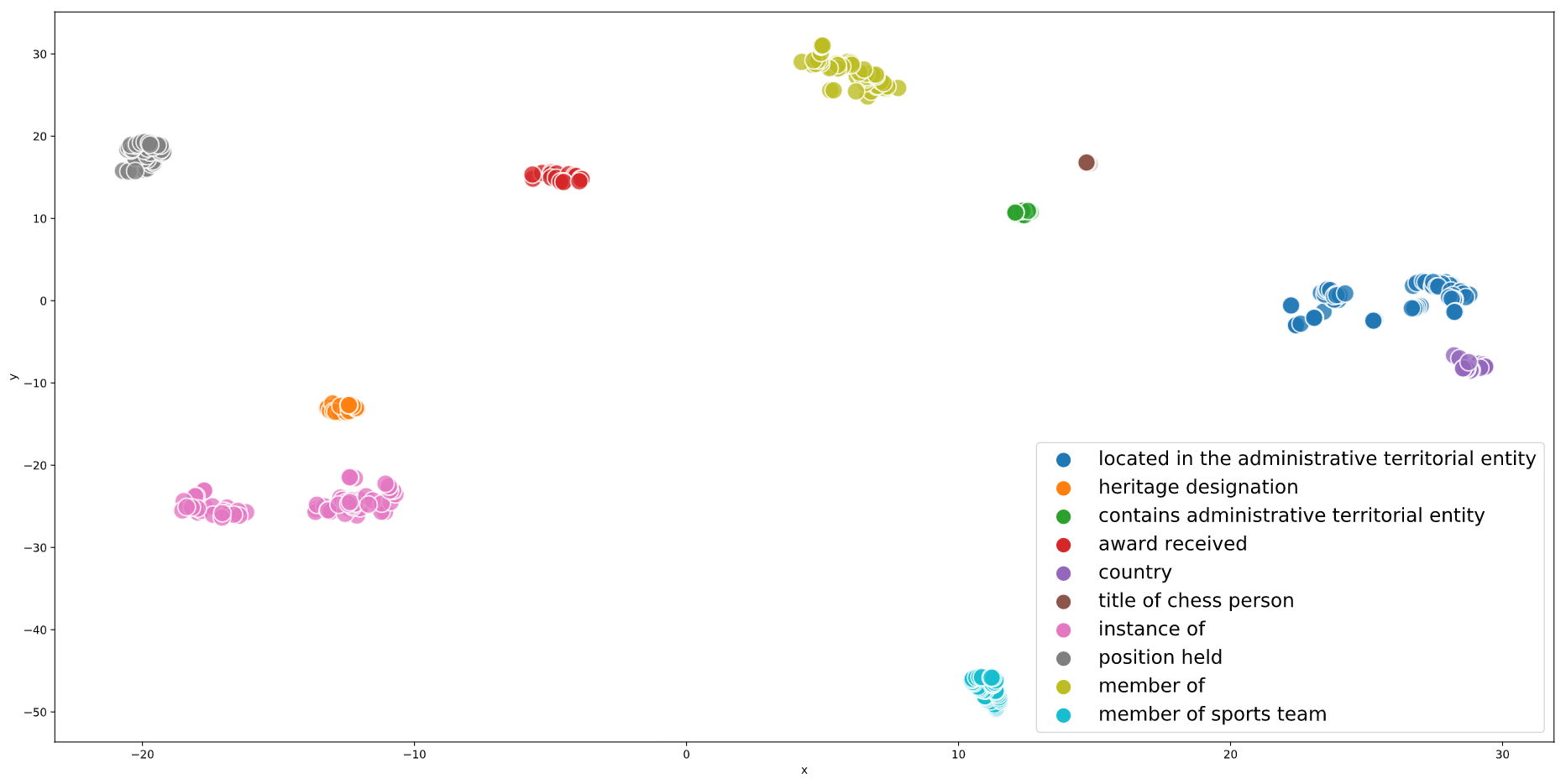}
    \caption{Two-dimensional t-SNE projection of the embedding learned by the model. For legibility, only the ten most common predicates are plotted.}
    \label{fig:t-sne_wikidata_merge}
\end{figure}

Specifically, we use the embeddings of a model learned on the Wikidata12k dataset transformed with the \splime\ merge method with a shrink factor of 4. The accompanying t-SNE plot is displayed in Figure~\ref{fig:t-sne_wikidata_merge}. We observe that predicates of the same type are mostly clustered together in the embedding space. While we create the t-SNE plot for all predicates, only the ten most common predicates are plotted for legibility. Otherwise, there would be some  predicates scattered throughout. This is most true for the predicates to which few splits have been applied, for example \textit{winner of an event (P1346)}, which was only split once.

\begin{figure}
    \begin{subfigure}{\textwidth}
        \centering
        \includegraphics[width=\textwidth]{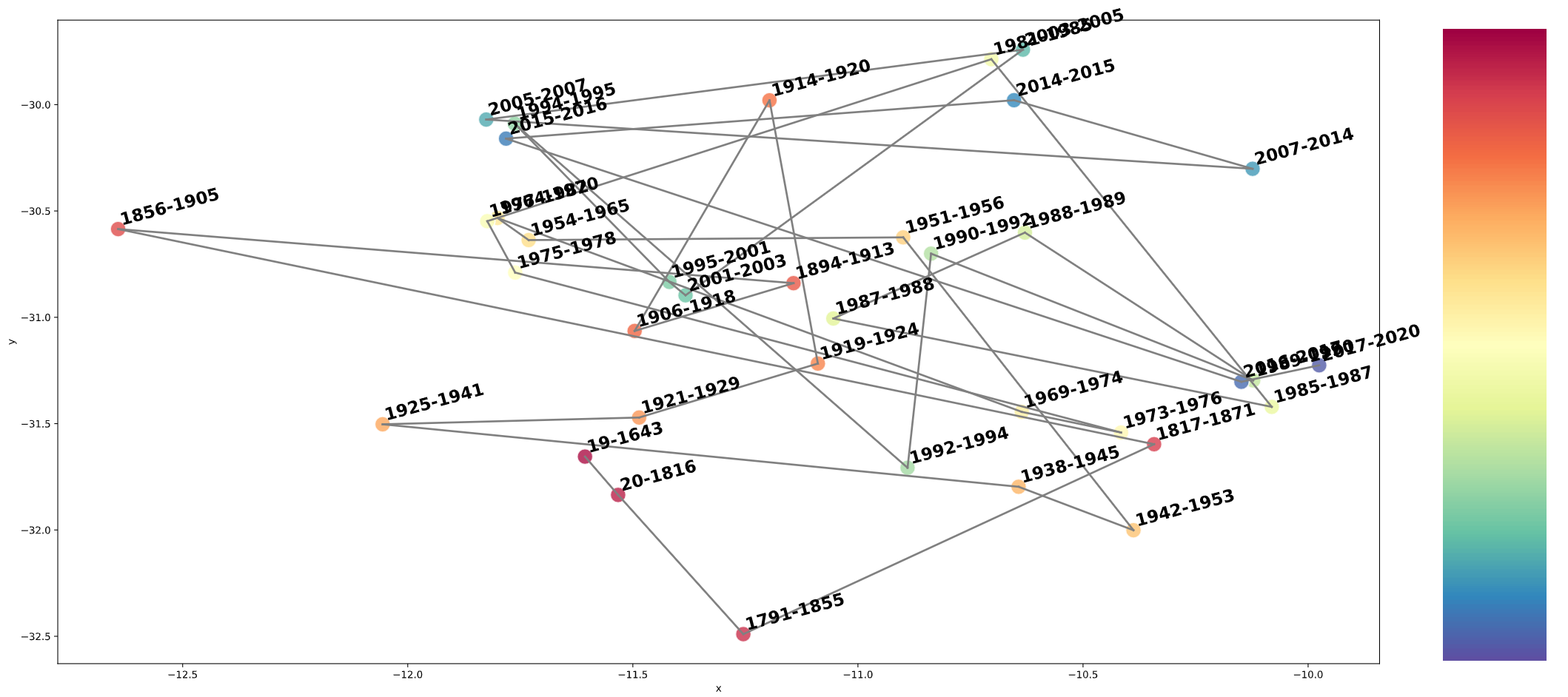}
        \caption{t-SNE plot of a dataset created by applying splits randomly. Only the \textit{residence (P551)} predicate is drawn.}
        \label{fig:t-sne_wikidata_random_p551}
    \end{subfigure}
    \vspace*{1 cm}
    \newline
    \begin{subfigure}{\textwidth}
        \centering
        \includegraphics[width=\textwidth]{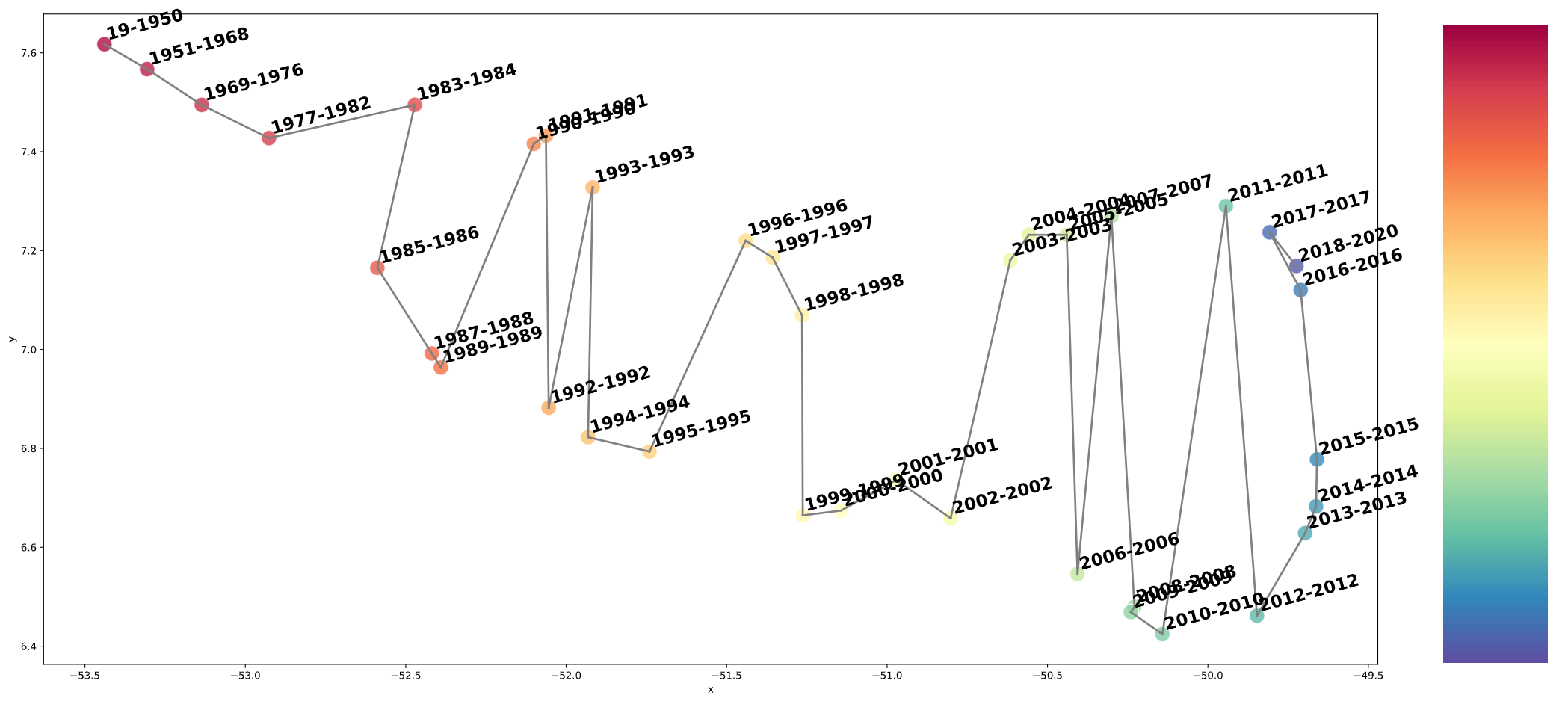}
        \caption{t-SNE plot of a dataset created using the \splime\ merge method. Only the \textit{member of sports team (P54)} predicate is drawn.  }
        \label{fig:t-sne_wikidata_merge_p54}
    \end{subfigure}
    \caption{Two t-SNE projections. The colors are based on the end-time of the predicate, ranging from red (long ago) to blue (recent). The lines between the points illustrate how the embedding moves through the vector space, and the labels denote the time period during which the predicate is valid.}
\end{figure}

Additionally, we note that predicates with many points seem to form lines or circles in the reduced dimensionality plot. To investigate this, we took the same t-SNE embeddings and created a plot containing just the predicate \textit{member of sports team (P54)}. This plot is displayed in Figure~\ref{fig:t-sne_wikidata_merge_p54}. It shows that the model has successfully learned a somewhat smooth temporal evolution of the embedding: each step in time moves the embedding in a similar direction, and the different points in time are well separated.

To evaluate how much of this is due to \splime , we also plot a baseline dataset obtained by applying random splits, as described in Section~\ref{sec:baseline}, for a predicate that had an approximately equal number of splits. The original could not be used as none of our random baseline models had the required number of splits for that predicate. Figure~\ref{fig:t-sne_wikidata_random_p551} shows that here the temporal evolution of the embedding is completely erratic. This implies that \splime\ includes temporal scopes in an intelligent manner.

\section{Change Point Detection in SpliMe}\label{sec:cpdcost}
In this section we give a more detailed explanation of CPD and how it relates to \splime . Specifically, we will discuss the cost function and kernel method as used by \splime\ during change point detection. For a complete overview of kernel functions in CPD we also refer to \cite{garreau2017change}. 

Broadly speaking, CPD algorithms can be divided into two categories: online and offline. Online CPD analyses a time series of unknown (possibly infinite) length. The decision on whether a change point has occurred is made every time a new data point arrives. In contrast, offline CPD considers the entire data set at once and can thus look back in time to see when a change occurred. For \splime , the length of the time series is already known. Therefore, we are only interested in offline CPD. A survey of offline CPD algorithms can be found in \cite{truong2020selective}.

As noted earlier, we employ the typology defined in \cite{truong2020selective}. That is, CPD is a model selection problem which consists of selecting the "best" possible segmentation of a time series. A CPD algorithm has three components a cost function, a search method and a constraint (on the number of change points). The cost function is a measure of homogeneity, i.e., how similar data points in a given segment are. The search method concerns locating possible segment boundaries, i.e., it locates parts of the signal that should be grouped together. The best possible segmentation is the one that minimizes the associated cost function.

Let $S = \{x_1,x_2,\dots, x_l\}$ denote a time series, and $L = \{k_1, \ldots, k_K\}$, $0 \leq K \leq l \in \mathbb{N}^{+}$ denote a possible segmentation of that time series. As explained in Section~\ref{sec:cpdsplit}, \splime\ creates a time series for every predicate in the graph. For each such time series $S_r \ r \in \predSet$, its entries $x_i, \ldots, x_l$ are vectors containing the proximity scores between pairs of nodes in the graph. These scores are calculated using only edges consisting of the predicate $r$ which exist at the given timestamp.

Because these vectors signify proximity scores, the data does not necessarily occupy a euclidean space.
As a result, we cannot use tradition distance metrics such as the $\ell_1$ and $\ell_2$ norms to calculate the distance between two samples.
Instead, a kernel function is used instead to map the data to a different space in which a comparison can be made. To do this, \splime\ uses the well known radial basis function kernel, defined as 
$$ \mathit{rbf}(x_1, x_2) = \exp(-\gamma || x_1 - x_2 ||^2)$$
Here, $x_1$ and $x_2$ are the two samples from $S_r$ being evaluated.
$||\cdot||^2$ is the $\ell_2$-norm and $\gamma > 0$ is the \textit{bandwidth parameter} determined according to the median heuristic.

As a cost function \splime\ uses kernelized mean change. Like the well known sum of squares metric, kernelized mean change calculates how far each sample lies from the (empirical) mean and sums the result. However, each sample is first translated with the $rbf$. Formally, the cost function for a subsection of the signal can be written as  
$$c(S_{a,b}) = \sum_{a \leq i < b} = ||\mathit{rbf}(S_i) - \overline{\mu}_{a,b}||^2 $$
where $\overline{\mu}_{a,b}$ denotes the empirical mean of the subsection in the transformed space. The total cost $cost(\cdot)$ for a segmentation $L$ then is the summation of the costs of each segment, i.e.,
$$ cost(L) = \sum_{i = 1}^{K - 1} c(S_{L_{i},L_{i+1}})$$

\section{Extended Related Work}\label{sec:extendedrelatedwork}
In this section we will provide an overview of several related static knowledge graph embedding works. A thorough list of static KG embedding models can be found in~\cite{ji2020survey,nickel2015review}. Firstly, as one of the earliest link prediction methods, Liben-Nowell et al. \cite{NowellLinkPrediction} apply network proximity measures to the problem of link prediction in homogenous graphs (e.g. social networks). Explained in detail in Section~\ref{sec:proximity}, network proximity measures calculate the proximity between a pair of nodes in a graph based on the graphs structure. Specifically, their measures calculate proximity scores for (a selection of) pairs of nodes in a graph. The pairs with the highest scores are the most similar and are intuitively the most likely to form a new link. 

\subsection{Representation Learning}\label{representationlearning}
Current state-of-the-art KG embedding methods perform representation learning. I.e., they learn a vector representation of the entities and predicates making up a KG. These vectors represent the \emph{latent features} of entities and relationships: the underlying parameters that determine their interactions. 

Roughly speaking, KG embedding approaches using latent features can be defined in two groups, \emph{translational} approaches and \emph{tensor-factorization} approaches. Models of the first type apply meaning to latent representation: entities that are similar must be close together according to some distance measure. Models of the second type do not apply any meaning to the embedding themselves, but capture the underlying interactions directly. We will now give an overview of several important KG embedding models, including an overview of some temporal KG embedding models.

\subsubsection{Translational Models} 
One of the most well known embedding models is \textit{TransE}~\cite{bordes2013translating}. It is based on the intuition that summing the subject and predicate embedding vectors should result in a vector approximately equal to the object embedding vector. Scores are assigned based on the $\ell_1$ or $\ell_2$ norms between the expected and actual object embedding. 

\textit{TransH}~\cite{wang2014knowledge} further increases the expressiveness of TransE by enabling each entity to have a unique representation for each predicate. This is achieved by modelling each predicate with two vectors. The first defines a hyperplane, the second defines the translation on that hyperplane. When calculating distance, the entity vectors are first projected on the predicate-specific hyperplane using the dot product. 

\subsubsection{Factorization models} One of the oldest factorization models is Canonical Polyadic (CP) decomposition. CP was long thought to be unsuitable for KGE because it learns separate representations for subject and object occurrences of an entity. However, \cite{NIPS2018_7682} propose a simple enhancement to CP that addresses this independence, called \textit{SimplE}. Specifically, they learn an additional inverse embedding for each predicate. The score of a triple is then calculated by taking the average between its normal score and its inverse score. 

ComplEx~\cite{pmlr-v48-trouillon16} observe that while the representation of an entity should be equal regardless of whether it occurs as a subject or as an object in a triple, its behavior should not be. Noting that the definition of the dot product on complex numbers is not symmetric, they suggest combining complex vectors with a dot product based scoring function. Now, the representation of an entity is the same, but its behavior depends on whether it is used as object or subject.

RotatE~\cite{sun2018rotate} models entities and predicates with complex vectors, and then views  relations as rotations from the subject to the object entity in the complex space. The authors go on to define several relationship patterns and prove that RotatE can, unlike its competitors, model all of these. 
\end{document}